\documentclass{article}
\usepackage{spconf,amsmath,graphicx,caption,tabularx,subcaption,censor,hyperref}
\usepackage[numbers]{natbib}

\title{Fast-BEV++: Fast by Algorithm, Deployable by Design}
\name{
\begin{tabular}[t]{c}
Yuanpeng Chen$^{1\ddagger \clubsuit}$, Hui Song$^{1\ddagger}$, Sheng Yang$^{2}$, Wei Tao$^1$, Shanhui Mo$^3$, \\ 
Shuang Zhang$^1$, Xiao Hua$^1$, Tiankun Zhao$^1$ \\
\small $\ddagger$ \textup{Equal Contribution} \quad $\clubsuit$ \textup{Project Leader}
\end{tabular}
\vspace{-2ex}
}
\address{
$^1$iMotion Automotive Technology (Suzhou) Co., Ltd \\
$^2$School of Data Science, Fudan University \quad $^3$Independent Researcher \\
\small Project Page: \url{https://github.com/ymlab/advanced-fastbev}
}
\begin{document}
%
\maketitle
\begin{abstract}
The advancement of vision-only BEV (Bird's-Eye-View) perception is hindered by the fundamental trade-off between perception accuracy and deployment efficiency. We introduce Fast-BEV++, resolving this tension through two principles: \textit{Fast by Algorithm} and \textit{Deployable by Design}. By decomposing view transformation into a hardware-oriented Index-Gather-Reshape pipeline, Fast-BEV++ eliminates custom kernels while achieving no less than $3\times$ speedup over baseline methods. Empirically, Fast-BEV++ establishes a new \textit{state-of-the-art} accuracy-speed trade-off on nuScenes, achieving 0.488 NDS while sustaining real-time inference at over 134 FPS. In particular, depth supervision yields consistent and tangible performance gains, maintaining the highest accuracy among comparable methods. The decomposed architecture enables seamless real-time deployment on production-level platforms, eliminating hardware constraints without loss of efficiency. Code and models are released on the linked project page.
\end{abstract}
\begin{keywords}
Computer Vision, 3D Object Detection, BEV (Bird’s-Eye View) Representation, Efficient Inference, Autonomous Driving
\end{keywords}

\section{Introduction}
\label{sec:intro}

The evolution of autonomous driving is increasingly converging toward pure-vision BEV perception~\citep{Jonah2020lss, li2022bevformer, li2022bevdepth}. This paradigm maps multi-camera image features into a unified ego-vehicle-centric BEV space, delivering a low-cost, semantically compact and cross-view consistent representation. It serves as a universal foundation for core downstream tasks including 3D object detection, semantic map segmentation~\citep{liu2022bevfusion} and motion planning~\citep{Hu2022uniad}, and has thus become a mainstream research direction for vision-centric autonomous driving.

Despite extensive methodological research and iterative development of pure-vision BEV perception, the field is still limited by a fundamental trade-off between perception accuracy and on-vehicle deployment viability. This bottleneck is rooted in mainstream view transformation designs: they either adopt computationally expensive operations that cannot satisfy automotive real-time constraints, or use platform-specific custom kernels that cripple cross-platform portability, which has become the core barrier to the mass production of BEV perception systems.

\begin{figure}[!t]
    \centering
    \includegraphics[width=0.5\textwidth]{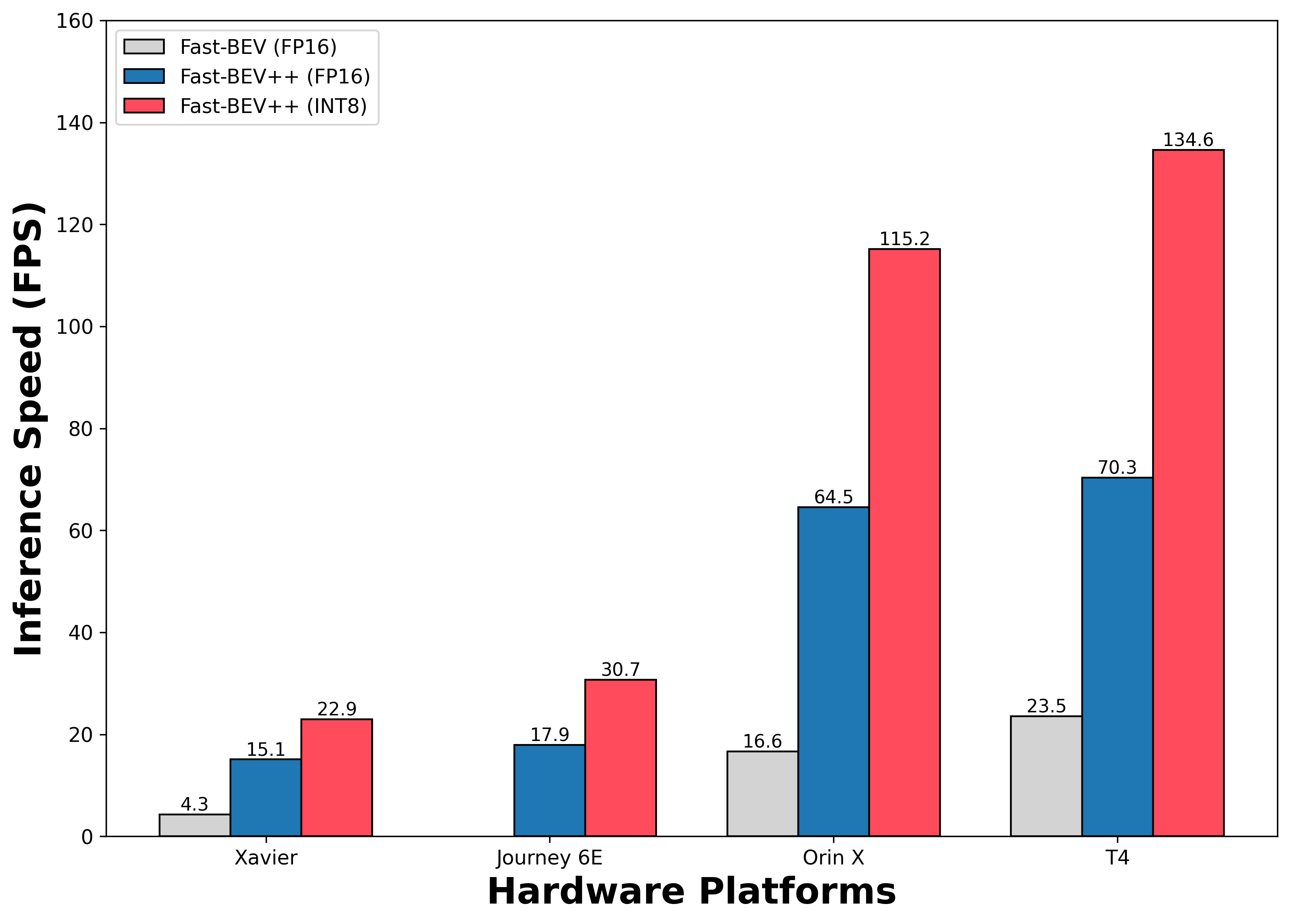}
    \caption{Inference speed of Fast-BEV and Fast-BEV++ (ResNet-50) on different hardware under FP16/INT8 precision. Under identical FP16 precision, Fast-BEV++ achieves $3.5\times$, $3.9\times$, and $3.0\times$ speedups over Fast-BEV on NVIDIA Xavier, Orin X, and T4, respectively.}
    \label{fig:deploy_compare}
\end{figure}

The Fast-BEV~\citep{li2023fast} framework partially addresses this fundamental challenge with its pre-computation paradigm: its core \textit{Fast-Ray transform} pre-computes the static 2D-to-3D geometric mapping and stores it as a lightweight Look-Up Table (LUT), effectively reducing inference latency. However, Fast-BEV’s monolithic design comes with inherent trade-offs: the strong coupling between static geometric priors and fixed LUT implementations limits architectural flexibility. This coupling presents non-trivial challenges for auxiliary perceptual cue integration, results in suboptimal memory efficiency, and leads to compromised cross-hardware portability, ultimately motivating our exploration of a truly hardware-agnostic view transformation paradigm.


\begin{figure*}[!htbp]
    \centering
    \begin{subfigure}{0.33\textwidth}
        \centering
        \includegraphics[width=\linewidth]{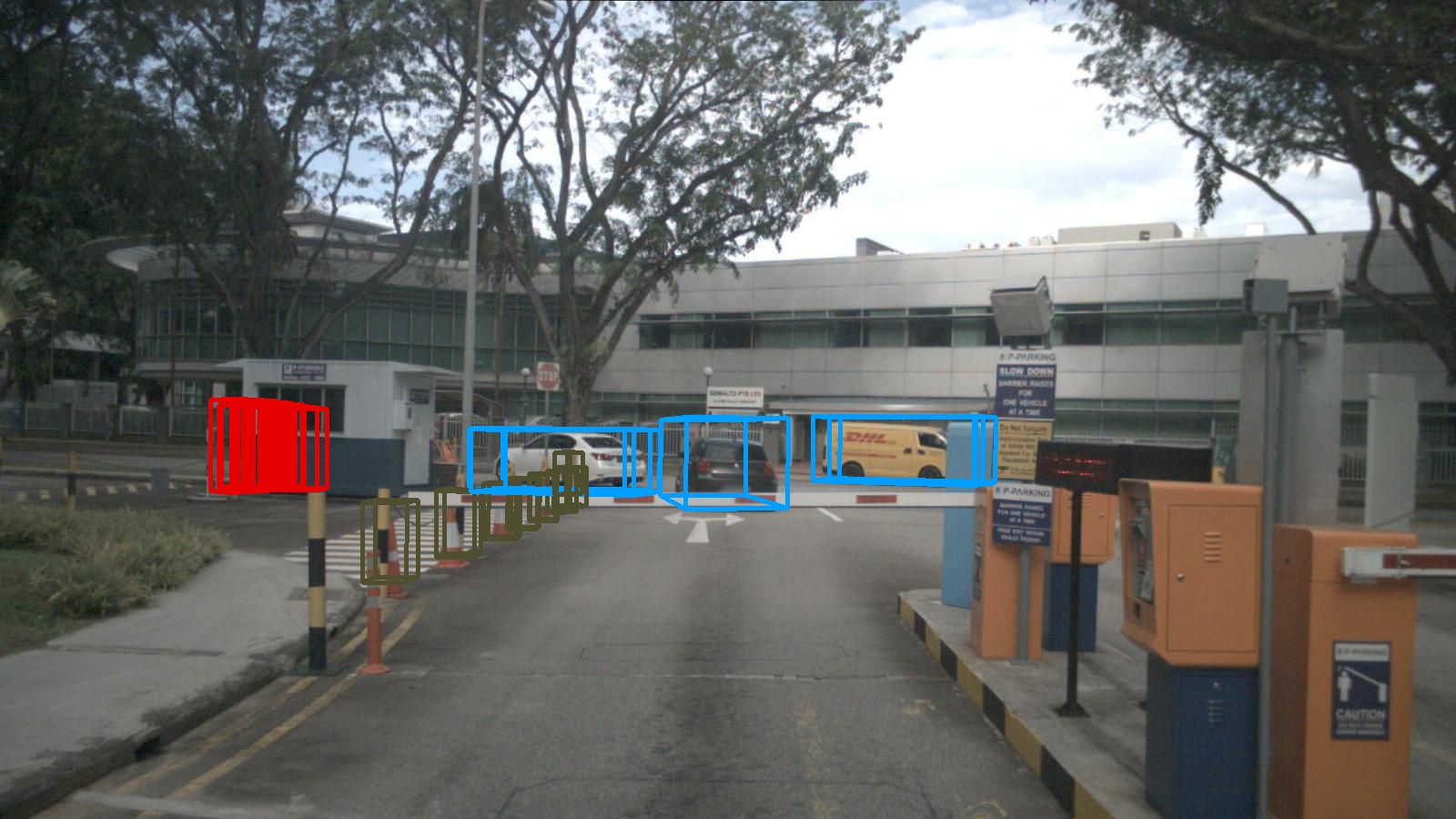}
        \caption{Front}
        \label{subfig:cam_front}
    \end{subfigure}
    \hfill
    \begin{subfigure}{0.33\textwidth}
        \centering
        \includegraphics[width=\linewidth]{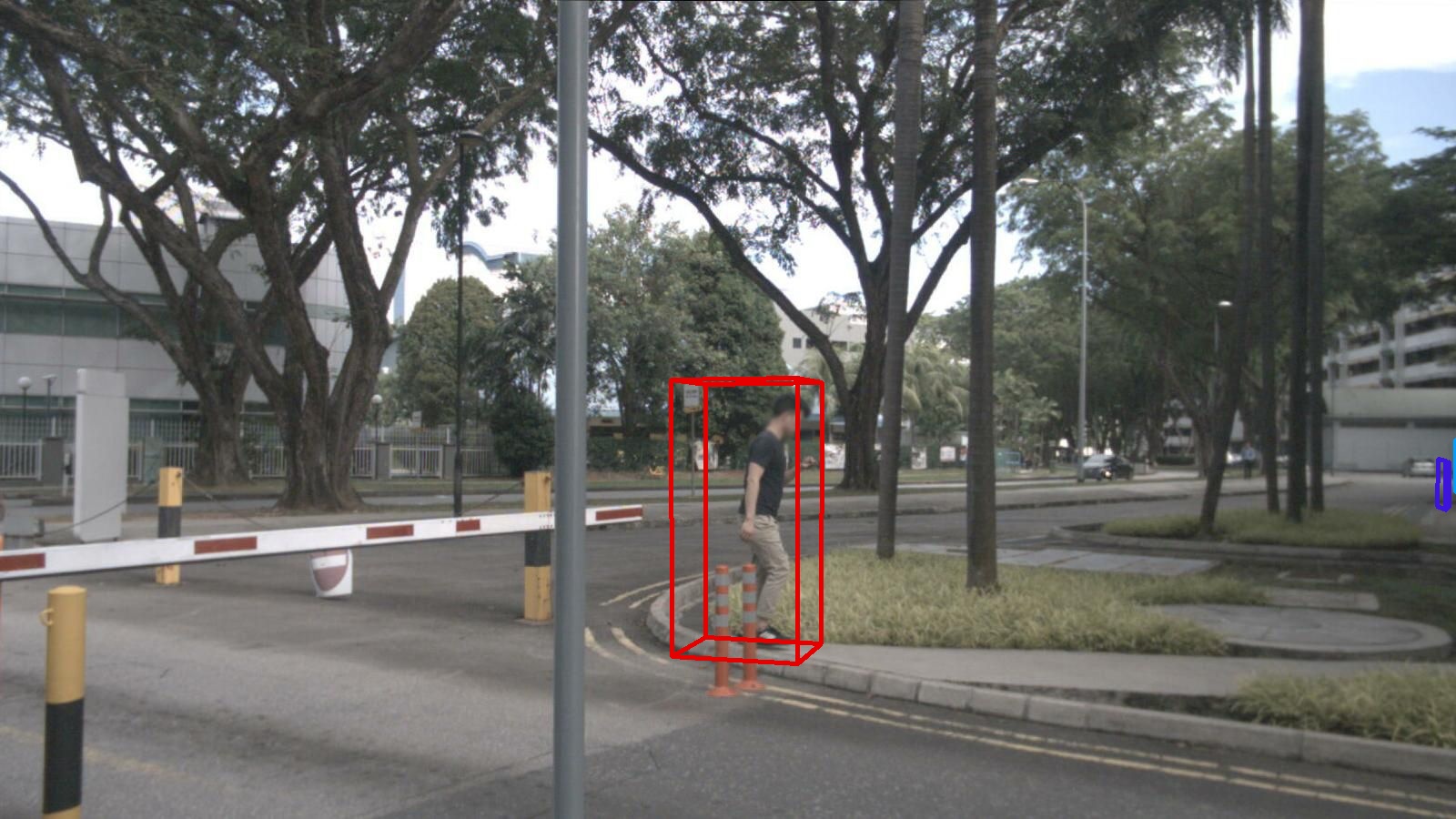}
        \caption{Front Right}
        \label{subfig:cam_front_right}
    \end{subfigure}
    \hfill
    \begin{subfigure}{0.33\textwidth}
        \centering
        \includegraphics[width=\linewidth]{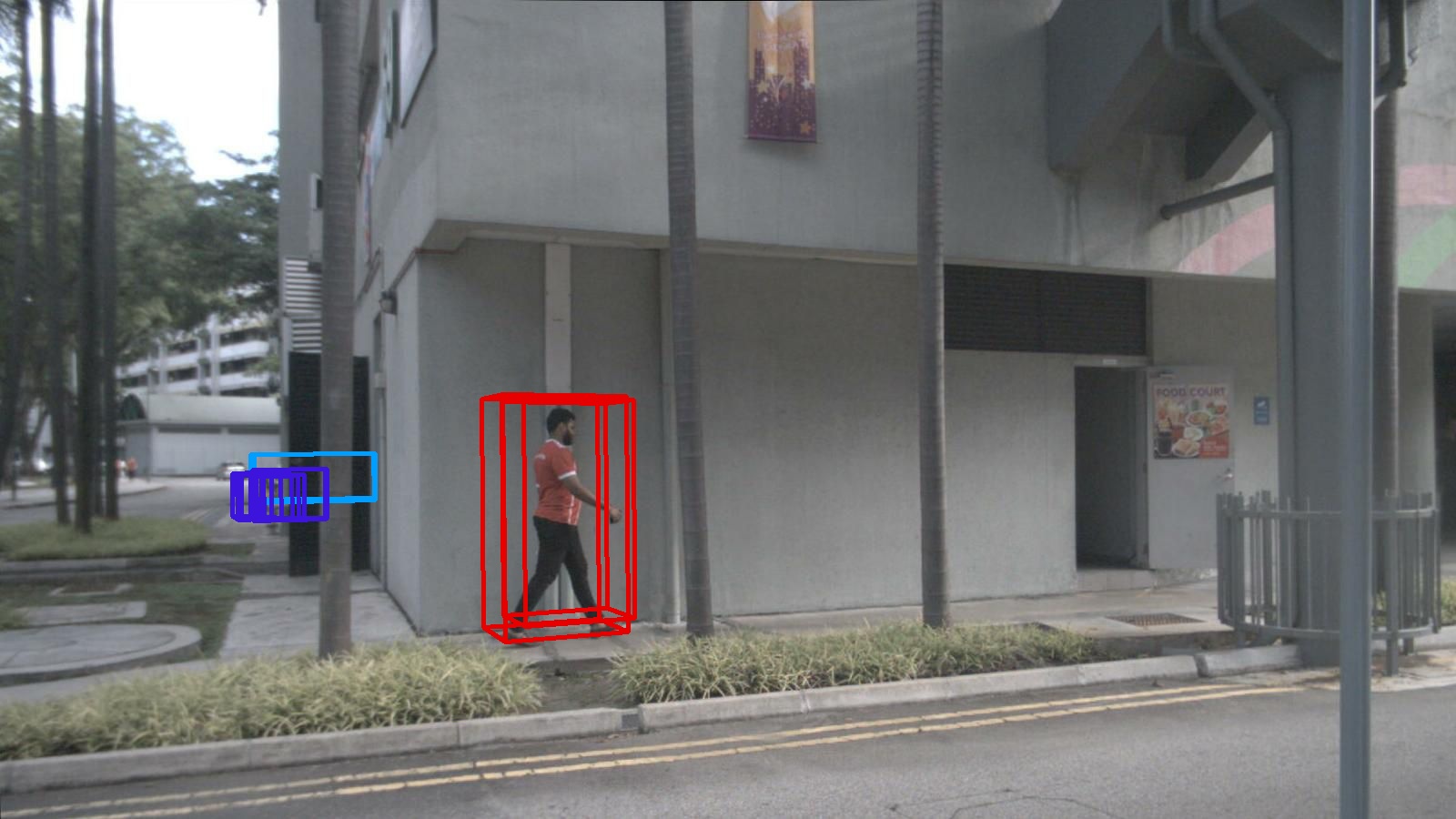}
        \caption{Back Right}
        \label{subfig:cam_back_right}
    \end{subfigure}

    \vspace{8pt}
    \begin{subfigure}{0.33\textwidth}
        \centering
        \includegraphics[width=\linewidth]{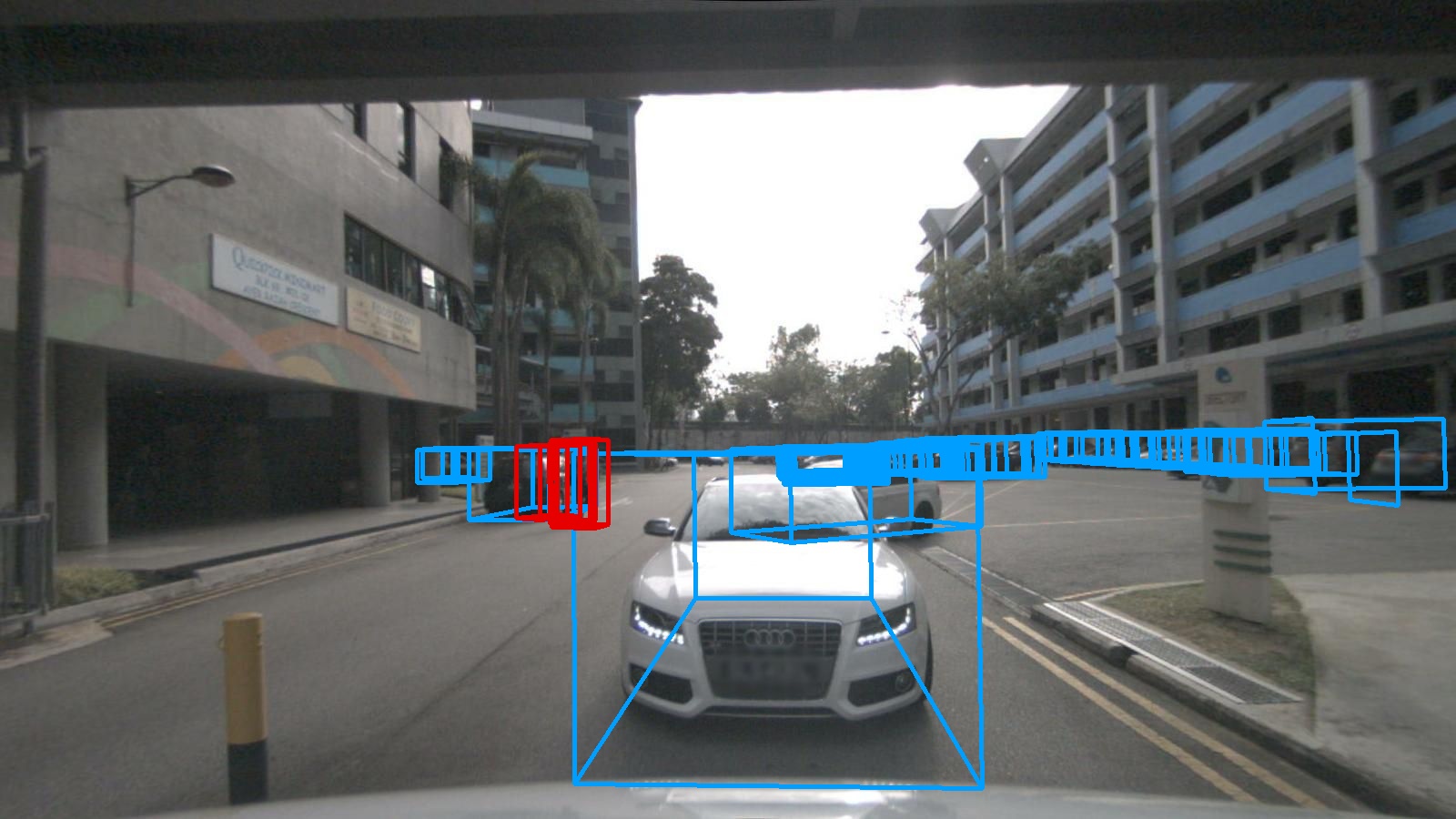}
        \caption{Back}
        \label{subfig:cam_back}
    \end{subfigure}
    \hfill
    \begin{subfigure}{0.33\textwidth}
        \centering
        \includegraphics[width=\linewidth]{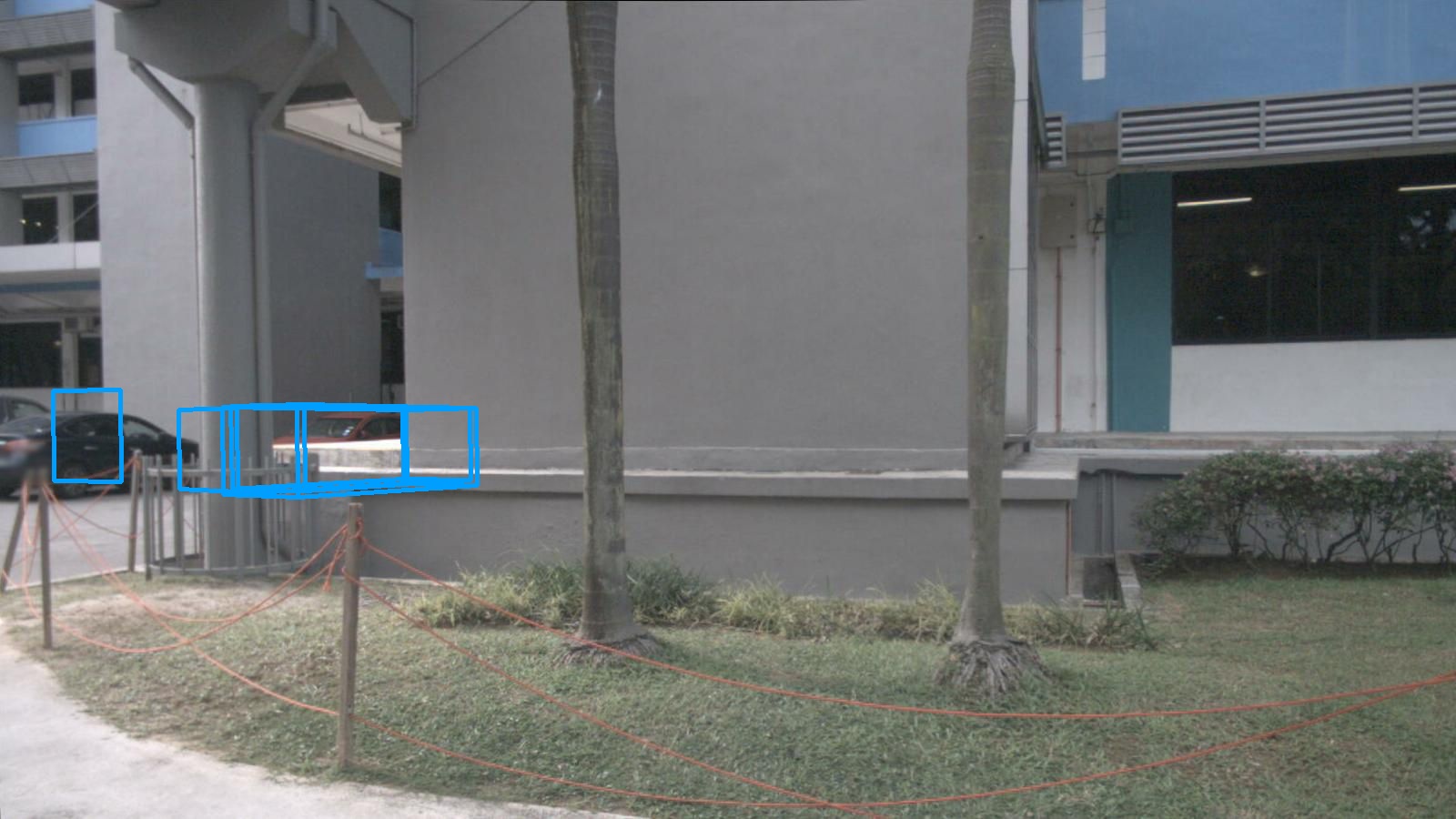}
        \caption{Back Left}
        \label{subfig:cam_back_left}
    \end{subfigure}
    \hfill
    \begin{subfigure}{0.33\textwidth}
        \centering
        \includegraphics[width=\linewidth]{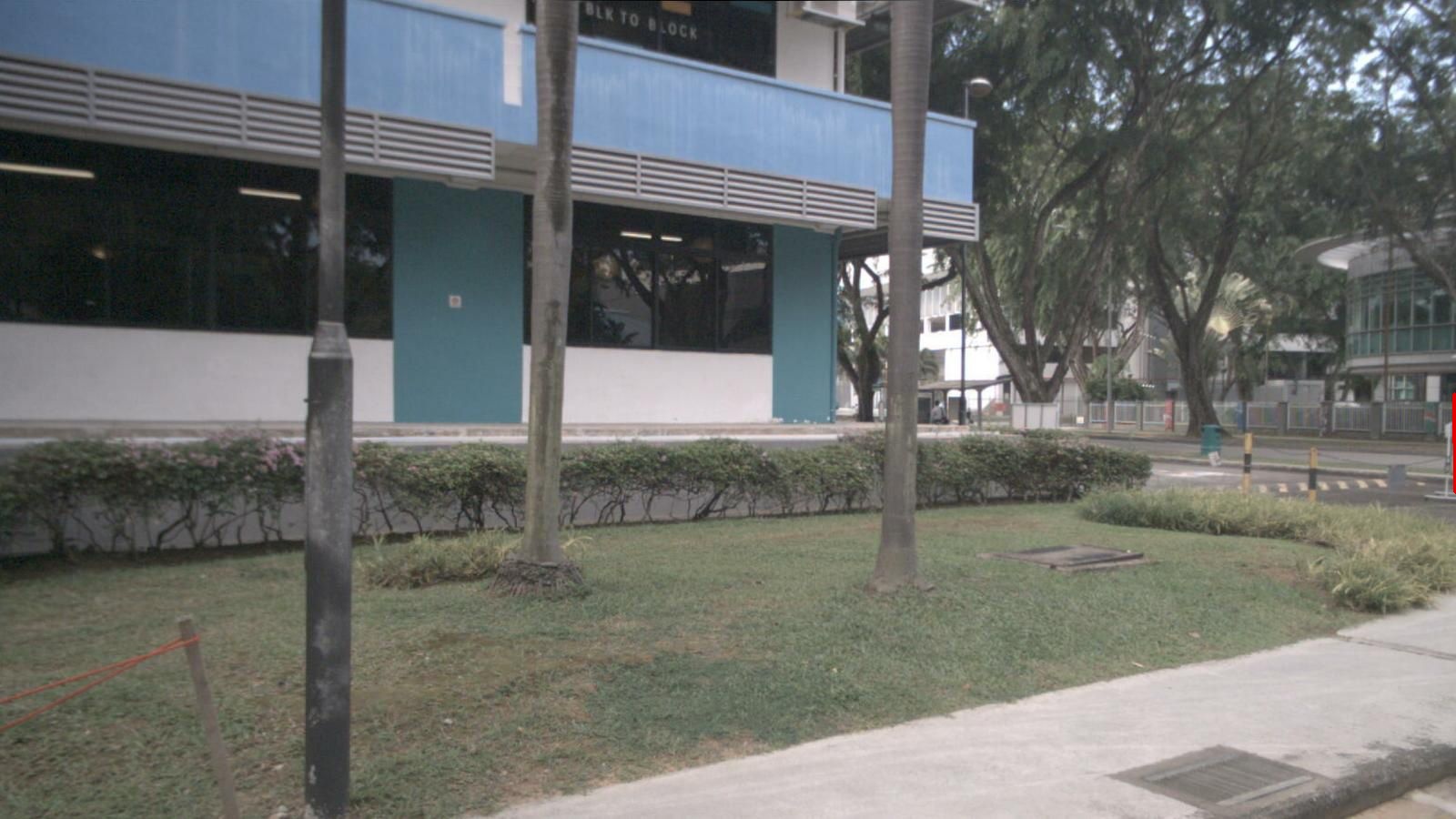}
        \caption{Front Left}
        \label{subfig:cam_front_left}
    \end{subfigure}

    \caption{Visualization of multi-view images from the nuScenes dataset in a typical complex urban scene, sorted in the standard clockwise order, where bounding boxes represent the outputs of Fast-BEV++ for typical traffic participants.}
    \label{fig:nuscenes_cameras}
\end{figure*}

In this paper, we propose Fast-BEV++, a framework inspired by Fast-BEV to deliver upgrades for deployment-oriented BEV perception. We demonstrate the aforementioned accuracy-deployment trade-off is effectively mitigated by reframing view transformation as standard, compiler-friendly tensor operations rather than a monolithic task-specific operation. Our key contributions are summarized as follows:

\begin{itemize}
    \item \textbf{We propose a view transformation methodology decomposed into a standard \textit{Index-Gather-Reshape} pipeline}, eliminating dependency on custom kernels and enabling a fully TensorRT-native implementation with zero custom plugins for BEV perception.
    \item \textbf{We demonstrate this decomposed architecture supports depth-aware fusion}: integrating learned depth priors directly into the feature gathering stage delivers significant performance gains without compromising deployment efficiency or inference latency.
    \item We empirically validate Fast-BEV++ achieves competitive performance on the nuScenes 3D object detection benchmark. Tests on production-grade automotive edge hardware further confirm optimizing for deployment tractability avoids sacrificing perception performance while delivering meaningful accuracy gains.
\end{itemize}

\section{Related Works}
\label{sec:related}

\noindent \textbf{Pure-Vision BEV Perception.} Robust 3D perception serves as a fundamental cornerstone of autonomous driving. While LiDAR-based methods such as VoxelNet~\citep{Y.Zhou2018Voxelnet}, SECOND~\citep{yan2018second}, PointRCNN~\citep{S.Shi2019Pointrcnn}, and PointPillars~\citep{Y.Zhou2019Pointpillars} have long dominated leading performance on mainstream 3D perception benchmarks, their prohibitive hardware cost and inherent sparse point cloud limitations pose critical barriers to large-scale deployment, driving the rapid rise of pure-vision BEV perception. Works including BEVDepth~\citep{li2022bevdepth}, DETR3D~\citep{Y.Wang2022Detr3d} have demonstrated the capability of this paradigm to generate semantically rich, spatially coherent 3D representations from widely available camera inputs, establishing a unified ego-centric BEV feature space as the standard foundation for downstream tasks such as 3D object detection.

\noindent \textbf{View Transformation.} View transformation serves as the key operation that projects multi-view 2D image features into a unified BEV representation. As an inherently ill-posed problem, it has been partially solved by deep learning models, yet existing methods still fall into two mainstream paradigms with inherent accuracy–deployment trade-offs.

The first is depth-based projection, pioneered by the LSS~\citep{Jonah2020lss, hu2023ealss} framework, which lifts 2D features to 3D space via pixel-wise depth estimation and aggregates these 3D features into a BEV grid. Methods such as BEVFusion~\citep{liu2022bevfusion} further enhance depth estimation with strong supervision, yet they still involve heavy computation and depend on specialized operators, limiting deployment on edge platforms.

The second is query-based aggregation, represented by BEVFormer~\citep{li2022bevformer, Yang2022BEVFormerVA}. It constructs BEV features via learnable queries and spatiotemporal attention without explicit depth prediction. However, attention’s quadratic complexity tends to bring high inference latency, limiting stable real-time inference on resource-constrained driving platforms. Similar problems exist in PETR~\citep{YingfeiLiu2022Petr, YingfeiLiu2022Petrv2}, DETR3D and FB-BEV~\citep{li2023fbbev}.

\noindent \textbf{Deployment-Oriented Optimization.} A parallel line of research addresses the fundamental performance-deployment trade-off for on-board autonomous driving applications, via deployment-friendly and hardware-efficient architectural designs. M$^2$BEV~\citep{EnzeXie2022M^2bev} laid the early foundational groundwork for efficient BEV feature extraction in multi-camera 3D perception, with streamlined view transformation pipelines that cut down redundant projection computations. MatrixVT~\citep{zhou2022matrixvt} further reformulates the complex view-to-BEV vision mapping process as pure, hardware-friendly matrix multiplication via \textit{Ring-and-Ray Decomposition}, largely reducing redundant computation overhead during inference. Meanwhile, Fast-BEV~\citep{li2023fast} introduces a novel deployment-optimized paradigm: its core \textit{Fast-Ray transform} pre-calculates the fixed static geometric mapping from image pixel coordinates to 3D BEV voxels, stores it as a lightweight Look-Up Table (LUT), and substantially reduces end-to-end inference latency via a direct multi-view to single-voxel projection scheme, enabling real-time high-fidelity BEV feature extraction on resource-constrained edge devices.

\section{Methodology}
\label{sec:method}

The core design philosophy of Fast-BEV++ is guided by two foundational principles: \textit{Fast by Algorithm} and \textit{Deployable by Design}. The model is carefully designed to fully exploit its performance and scalability. To address the fundamental trade-off between perception accuracy and on-vehicle deployment tractability, we reframe the traditional monolithic 2D-to-3D view transformation into a fully decomposed, hardware-agnostic computational pipeline.

This section elaborates on the architectural evolution, operator-native view transformation paradigm, and end-to-end depth-aware fusion mechanism of Fast-BEV++.

\subsection{Limitations of Monolithic Transformation}
As the key mechanism of Fast-BEV, \textit{Fast-Ray transformation} achieves large latency reduction via a precomputed static Look-Up Table (LUT), encapsulating the entire 2D-to-3D projection into a single opaque operation:
\begin{equation}
\mathbf{V}_{\mathrm{BEV}} = \Psi (\mathbf{F}_{\mathrm{img}}, \mathrm{LUT})
\end{equation}
Despite achieving outstanding performance, this monolithic design leads to two critical deployment bottlenecks:
\begin{itemize}
    \item \textbf{Memory Fragmentation:} The custom aggregation in Fast-BEV scatters multi-view features into 3D grids without organized write operations, leading to fragmented memory access patterns. This fragmentation incurs data movement overhead and necessitates atomic read-modify-write operations to resolve voxel collisions. Such inefficiencies create a memory-bandwidth bottleneck, restricting the overall inference throughput on modern platforms.
    \item \textbf{Architectural Rigidity:} The opaque operator $\Psi$ creates a hard coupling between spatial mapping and feature extraction. This lack of transparency leads to a \textit{hardware lock-in} effect, where the model relies on non-portable custom kernels instead of compiler-native primitives. Consequently, this structural inflexibility creates a barrier to augmenting the BEV representation with external geometric information, such as end-to-end depth distributions.
\end{itemize}

\subsection{Operator-Native View Transformation}
To remove the reliance on custom kernels and mitigate memory fragmentation, Fast-BEV++ radically reformulates the view transformation into a two-stage computational graph built solely from standard, operator-native primitives.

\subsubsection{Deterministic Index Generation}

Instead of a static and unorganized LUT, we formulate the geometric mapping as a structured, precomputed index system designed for sequential memory access. First, given the fixed camera geometry, we establish a raw geometric correspondence by performing inverse projection for each voxel in the ego-centric BEV space. To eliminate multi-view aggregation overhead and the high-latency atomic operations typical of traditional methods, we enforce a strict \textbf{one-to-one voxel-pixel mapping} for overlapping regions. By assigning a deterministic priority to a single camera source for each voxel, we ensure that each 3D position draws from a unique 2D source and effectively resolves write-conflicts at the source.

Most importantly, we introduce a \textbf{deterministic sorting strategy} to resolve the fragmented memory issue. We rearrange all valid mappings to align strictly with the continuous spatial memory layout of the target BEV tensor following the pre-defined order. This pre-sorting process effectively queues all memory operations and transforms the transformation into two synchronized 1D dense index tensors via a mapping function $\mathcal{M}(\mathbf{p}_v)$:
\begin{equation}
    \mathcal{M}(\mathbf{p}_v) \to
    \begin{cases}
    \mathcal{I}_{\mathrm{spatial}} = \left\{ (\text{Cam}_k, u_k, v_k) \right\}_{k=1}^N \\
    \mathcal{I}_{\mathrm{depth}} = \left\{ (\text{Cam}_k, u_k, v_k, d_k) \right\}_{k=1}^N
    \end{cases}
\end{equation}
where $\mathcal{I}_{\mathrm{spatial}}$ enables sequential gathering of semantic features and $\mathcal{I}_{\mathrm{depth}}$ provides a synchronized path for depth distribution. Aligning the index order with the physical memory layout enables hardware to stream features in a single, cache-friendly pass with zero fragmented latency.

\subsubsection{Index-Gather-Reshape Operation}

During inference, the 2D-to-3D transformation is executed entirely using standard compiler-friendly operators to effectively bypass the memory bottlenecks. 

\noindent \textbf{Native Gather with Modulation.} We utilize the standard \texttt{Gather} operator to perform feature extraction and depth fusion simultaneously. By querying the synchronized indices, we extract semantic features and depth probabilities, fusing them via element-wise multiplication as follows:
\begin{equation}
    \mathbf{F}_{\mathrm{flat}} = 
    \underbrace{\text{Gather}\left(\mathbf{F}_{\mathrm{img}}, \mathcal{I}_{\mathrm{spatial}}\right)}_{\text{Semantic Feature}}
    \odot
    \underbrace{\text{Gather}\left(\mathbf{D}_{\mathrm{dist}}, \mathcal{I}_{\mathrm{depth}}\right)}_{\text{Depth Distribution}}
\end{equation}
Because the indices were pre-sorted in previous stage, the output $\mathbf{F}_{\mathrm{flat}}$ is generated as a highly contiguous 1D memory buffer rather than a fragmented collection of data. This strictly sequential read-write pattern maximizes cache hit rates and completely eliminates the fragmented memory movement overhead and atomic operations required in the baseline. By using native primitives, the entire process benefits from the extreme optimization provided by mainstream inference engines like TensorRT.

\noindent \textbf{Zero-cost Reshape.} Finally, we reconstruct the dense 3D BEV feature volume using a standard \texttt{Reshape} operator:
\begin{equation}
    \mathbf{V}_{\mathrm{BEV}} = \text{Reshape}(\mathbf{F}_{\mathrm{flat}}, [Z, H, W, C])
\end{equation}
Since the element order in $\mathbf{F}_{\mathrm{flat}}$ is perfectly aligned with the target spatial dimensions $(Z, H, W, C)$ via our deterministic sorting strategy, this \texttt{Reshape} operation is zero-cost. It involves only modifying the tensor metadata to redefine dimensions, with no arithmetic computation or physical memory movement, as the data already resides in the target physical layout.

\subsection{End-to-End Depth-Aware Fusion}

The decomposed pipeline serves as a versatile computational foundation, enabling the integration of explicit 3D geometry with negligible impact on deployment efficiency. By transcending the inherent rigidity of monolithic architectures, this design facilitates the seamless embedding of learned depth priors into the view transformation process, effectively replacing the uniform depth assumption. 

Specifically, a lightweight pixel-wise depth prediction head could be integrated in parallel with the 2D image encoder to generate a distribution $\mathbf{D}_{\mathrm{dist}}$ over predefined depth bins. Crucially, as the entire \texttt{Index}-\texttt{Gather}-\texttt{Reshape} pipeline is constructed from standard differentiable primitives, supervisory signals from the 3D detection head can propagate back to the 2D depth network without obstruction. This synergy allows for end-to-end joint optimization, ensuring that depth estimation is directly aligned with the downstream 3D perception task. Furthermore, since depth modulation is fused into the gathering stage via element-wise operations, this significant accuracy gain is achieved with nearly zero extra inference overhead.

\begin{table*}[!t]
    \centering
    \small
    \renewcommand{\arraystretch}{1.1}
    \setlength{\doublerulesep}{0.8pt}
    \caption{Comparison on the nuScenes \textit{val} set. All models adopt one of ResNet~\citep{He2015res}, ResNeXt~\citep{Xie2016resnext} (referred to as X), or Swin Transformer~\citep{liu2021Swin} as their backbones. ``L'', ``C'', and ``D'' denote LiDAR input, pure-vision (camera) input, and depth supervision, respectively; ``$\ddagger$'' indicates our method equipped with scale NMS and test-time augmentation.}
    \label{tab:nuscenes_comparison}
    \begin{tabular}{|l|c|c|c|c|c|c|c|c|}
        \hline
        \noalign{\hrule height 0.8pt}
        \textbf{Methods} & \textbf{Backbone} & \textbf{Image Res.} & \textbf{Modality} & \textbf{mAP}$\uparrow$ & \textbf{mATE}$\downarrow$ & \textbf{mASE}$\downarrow$ & \textbf{mAOE}$\downarrow$ & \textbf{NDS}$\uparrow$ \\
        \hline\hline
        CenterPoint-Pillar~\citep{TianweiYin2021CenterPoint} & - & - & L & 0.503 & - & - & - & 0.602 \\
        CenterPoint-Voxel~\citep{TianweiYin2021CenterPoint} & - & - & L & 0.564 & - & - & - & 0.648 \\
        \noalign{\hrule height 0.6pt}
        \hline
        FCOS3D~\citep{TaiWang2021Fcos3d} & R50 & 900$\times$1600 & C & 0.295 & 0.806 & 0.268 & 0.511 & 0.372 \\
        BEVDet~\citep{huang2021bevdet} & R50 & 256$\times$704  & C & 0.286 & 0.724 & 0.278 & 0.590 & 0.372 \\
        PETR~\citep{YingfeiLiu2022Petr} & R50 & 384$\times$1056 & C & 0.313 & 0.768 & 0.278 & 0.564 & 0.381 \\
        PETR~\citep{YingfeiLiu2022Petr} & Swin-Tiny & 512$\times$1408 & C & 0.361 & 0.732 & 0.273 & 0.497 & 0.431 \\
        BEVDet4D~\citep{huang2022bevdet4d} & Swin-Tiny & 256$\times$704  & C & 0.323 & 0.674 & 0.272 & 0.503 & 0.453 \\
        BEVDepth~\citep{li2022bevdepth} & R50 & 256$\times$704  & C\&D & 0.351 & 0.639 & 0.267 & 0.479 & 0.475 \\
        \hline
        Fast-BEV~\citep{li2023fast} & R50 & 256$\times$704 & C & 0.334 & 0.665 & 0.285 & 0.393 & 0.473 \\
        Fast-BEV$^\ddagger$~\citep{li2023fast} & R50 & 256$\times$704 & C & \textbf{0.346} & 0.667 & 0.285 & 0.401 & \textbf{0.477} \\
        \hline
        \textbf{Fast-BEV++} & R50 & 256$\times$704 & C & 0.344 & 0.656 & 0.285 & 0.512 & 0.478 \\
        \textbf{Fast-BEV++} & R50 & 256$\times$704 & C\&D & \textbf{0.359} & 0.728 & 0.281 & 0.581 & \textbf{0.488} \\
        \noalign{\hrule height 0.6pt}
        \hline
        FCOS3D~\citep{TaiWang2021Fcos3d} & R101 & 900$\times$1600 & C & 0.321 & 0.754 & 0.260 & 0.486 & 0.395 \\
        BEVDet~\citep{huang2021bevdet} & R101 & 512$\times$1408 & C & 0.349 & 0.637 & 0.269 & 0.490 & 0.417 \\
        PETR~\citep{YingfeiLiu2022Petr} & R101 & 512$\times$1408 & C & 0.357 & 0.710 & 0.270 & 0.490 & 0.421 \\
        DETR3D~\citep{Y.Wang2022Detr3d} & R101 & 900$\times$1600 & C & 0.347 & 0.765 & 0.267 & 0.392 & 0.422 \\
        MatrixVT~\citep{zhou2022matrixvt} & R101 & 512$\times$1408 & C & 0.396 & 0.577 & 0.261 & 0.397 & 0.467 \\
        M$^2$BEV~\citep{EnzeXie2022M^2bev} & X101 & 900$\times$1600 & C & 0.417 & 0.647 & 0.275 & 0.377 & 0.470 \\
        BEVDet4D~\citep{huang2022bevdet4d} & R101 & 640$\times$1600 & C & 0.396 & 0.619 & 0.260 & 0.361 & 0.515 \\
        BEVFormer~\citep{li2022bevformer} & R101 & 900$\times$1600 & C & 0.416 & 0.673 & 0.274 & 0.372 & 0.517 \\
        BEVDepth~\citep{li2022bevdepth} & R101 & 512$\times$1408 & C\&D & 0.412 & 0.565 & 0.266 & 0.358 & \textbf{0.535} \\
        \hline
        Fast-BEV~\citep{li2023fast} & R101 & 900$\times$1600 & C & 0.402 & 0.582 & 0.278 & 0.304 & 0.531 \\
        Fast-BEV$^\ddagger$~\citep{li2023fast} & R101 & 900$\times$1600 & C & \textbf{0.413} & 0.584 & 0.279 & 0.311 & \textbf{0.535} \\
        \hline
        \textbf{Fast-BEV++} & R101 & 896$\times$1600 & C & 0.397 & 0.633 & 0.284 & 0.513 & 0.507 \\
        \textbf{Fast-BEV++} & R101 & 896$\times$1600 & C\&D & \textbf{0.414} & 0.603 & 0.276 & 0.501 & \textbf{0.522} \\
        \noalign{\hrule height 0.8pt}
        \hline
    \end{tabular}
\end{table*}

\section{Experiments}
\label{sec:experiments}

\subsection{Experimental Setup}
\noindent \textbf{Datasets.}  All experiments are conducted on the nuScenes dataset, with training and evaluation strictly following the official training and validation splits.

\noindent \textbf{Metrics.} 
We use the official standard nuScenes 3D object detection metrics: mean Average Precision (mAP) for quantifying detection accuracy, mean Average Translation, Scale and Orientation Errors (mATE, mASE, mAOE) for measuring the translation, scale and orientation errors of predicted 3D bounding boxes, and nuScenes Detection Score (NDS) as the integrated comprehensive metric for overall detection performance evaluation.

\noindent \textbf{Training Details.} All models are trained on a computing node equipped with $8\times$ NVIDIA H20 (96GB) GPUs, with a total batch size of 64.  We train the models for 20 epochs using the AdamW~\citep{Loshchilov2017AdamW} optimizer ($\text{LR}=2 \times 10^{-4}$, $\text{WD}=1 \times 10^{-2}$), along with a warm-up and step decay learning rate schedule. The CBGS~\citep{Zhu2019CBGS} sampling strategy is adopted during training.

\noindent \textbf{Inference Details.} Inference validation is performed on the NVIDIA Jetson AGX Xavier, Horizon Robotics Journey 6E, NVIDIA Jetson AGX Orin X, and NVIDIA Tesla T4, all supporting both INT8 post-training quantization (PTQ), which substantially boosts inference efficiency by leveraging low-bit representation, leading to lower FLOPs and reduced memory bandwidth overhead, and FP16 precision.

\subsection{Main Results}
We empirically validate our framework on the nuScenes \textit{val} set against \textit{state-of-the-art} methods, as shown in Table~\ref{tab:nuscenes_comparison}. Our results demonstrate that Fast-BEV++ achieves exceptional performance.

Our lightweight (R50) version, adopting a camera-only approach, achieves a highly competitive NDS score of 0.478, outperforming the original Fast-BEV (0.477 NDS) as well as the strong depth-based baseline BEVDepth (R50) (0.475 NDS). This empirical result validates that our decomposed, and deployment-efficient view transformation module inherently confers a performance gain for visual perception. Building on this foundation, activating our end-to-end depth-aware fusion mechanism (C\&D) further elevates the model’s performance to 0.359 mAP and 0.488 NDS, representing a substantial improvement over all baseline methods with comparable architectural complexity.

Furthermore, our heavyweight (R101) version also performs on par with leading methods. Given our core optimizations primarily focus on inference acceleration and deployment efficiency, this result demonstrates its favorable comprehensive performance.

\subsection{Deployment Performance and Inference Speed}
To empirically validate our \textit{Deployable by Design} philosophy, we benchmark the end-to-end inference speed (measured in FPS) of the Fast-BEV++ (R50) model across a representative set of platforms, as shown in Figure~\ref{fig:deploy_compare}. We conduct evaluations under both FP16 and INT8 quantization settings.

By constructing the view transformation with compiler-friendly primitives, Fast-BEV++ fully unlocks the acceleration potential of modern parallel computing hardware. Notably, our method, especially the Fast-BEV++ (R50-CBGS) variant, establishes a new \textit{state-of-the-art} trade-off between accuracy and efficiency, achieving 134 FPS on a Tesla T4 GPU with INT8 quantization while maintaining 0.489 NDS. This validates its practical readiness for real-world, production-level deployment without relying on custom operators.

The results summarize two key observations:
\begin{itemize}
    \item \textbf{Performance Scaling:} The framework shows linear scaling of inference throughput with hardware computational capacity, yielding proportional speedups across platforms.
    \item \textbf{Low-Precision Quantization:} INT8 quantization delivers consistent efficiency gains across all platforms. On mainstream platforms including NVIDIA Orin X and T4, inference speeds exceed 100 FPS.
\end{itemize}

\begin{figure*}[!htbp]
    \centering
    \begin{subfigure}{0.33\textwidth}
        \centering
        \includegraphics[width=\linewidth]{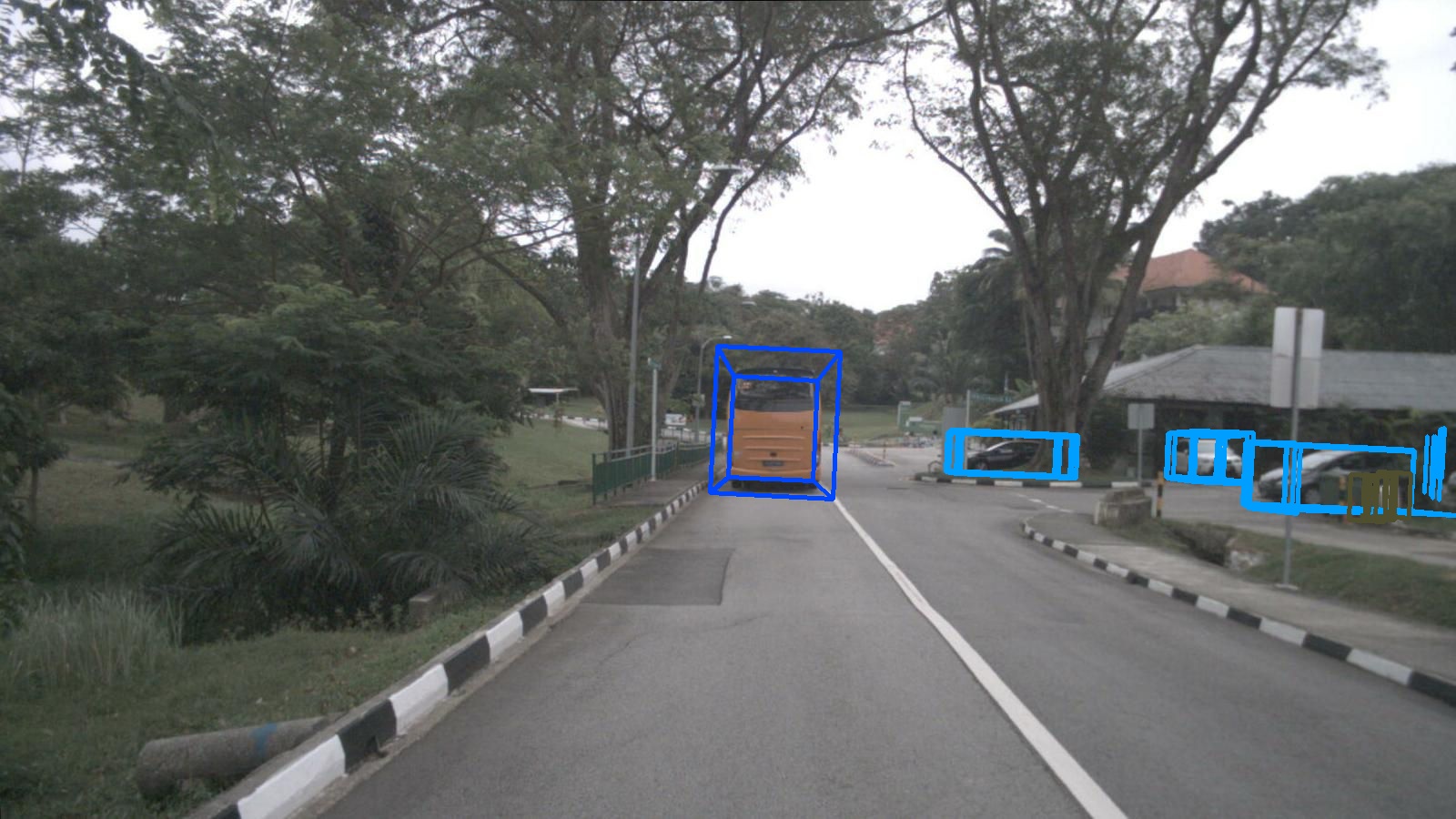}
        \caption{Front}
        \label{subfig:cam_front_more}
    \end{subfigure}
    \hfill
    \begin{subfigure}{0.33\textwidth}
        \centering
        \includegraphics[width=\linewidth]{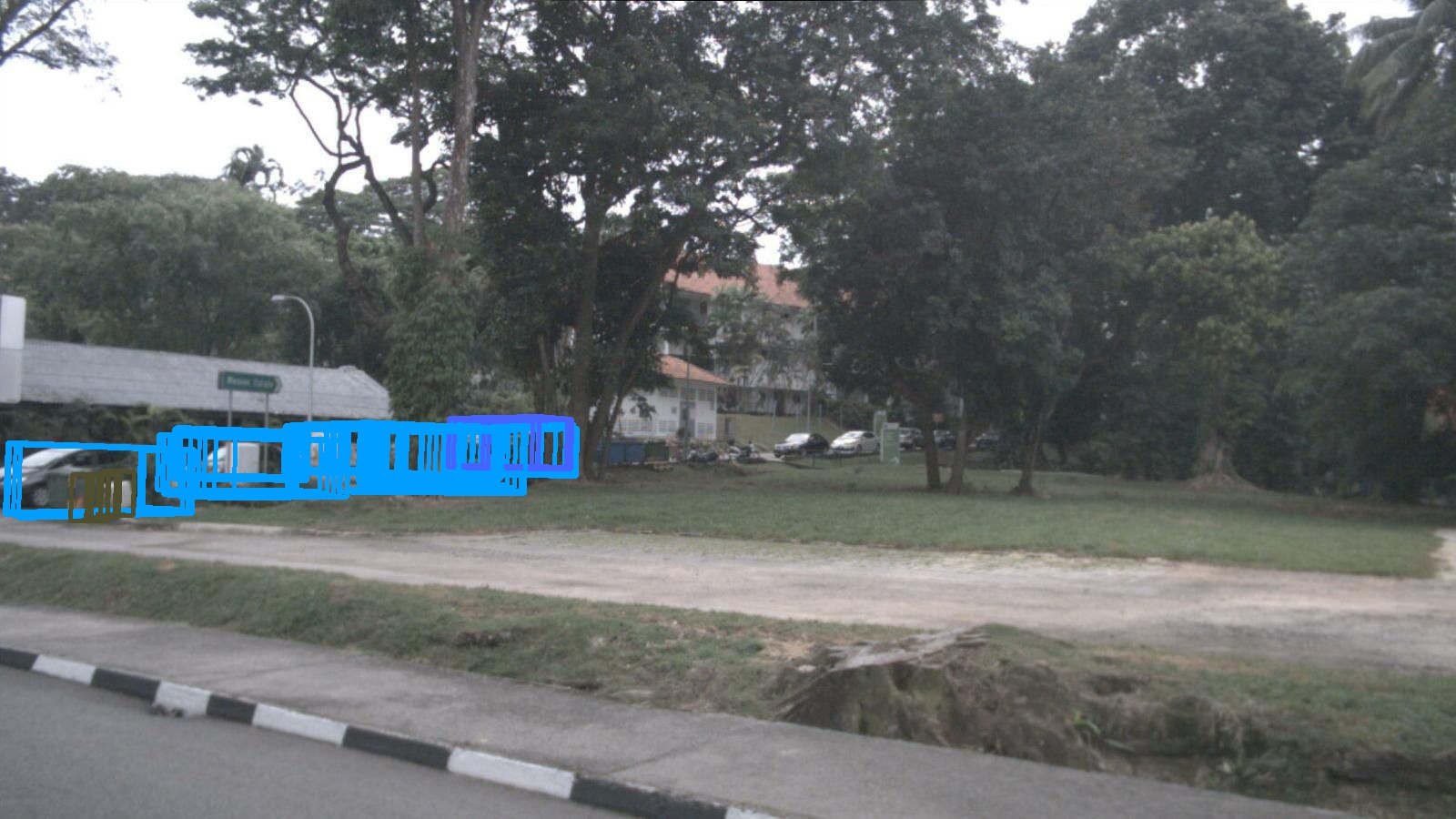}
        \caption{Front Right}
        \label{subfig:cam_front_right_more}
    \end{subfigure}
    \hfill
    \begin{subfigure}{0.33\textwidth}
        \centering
        \includegraphics[width=\linewidth]{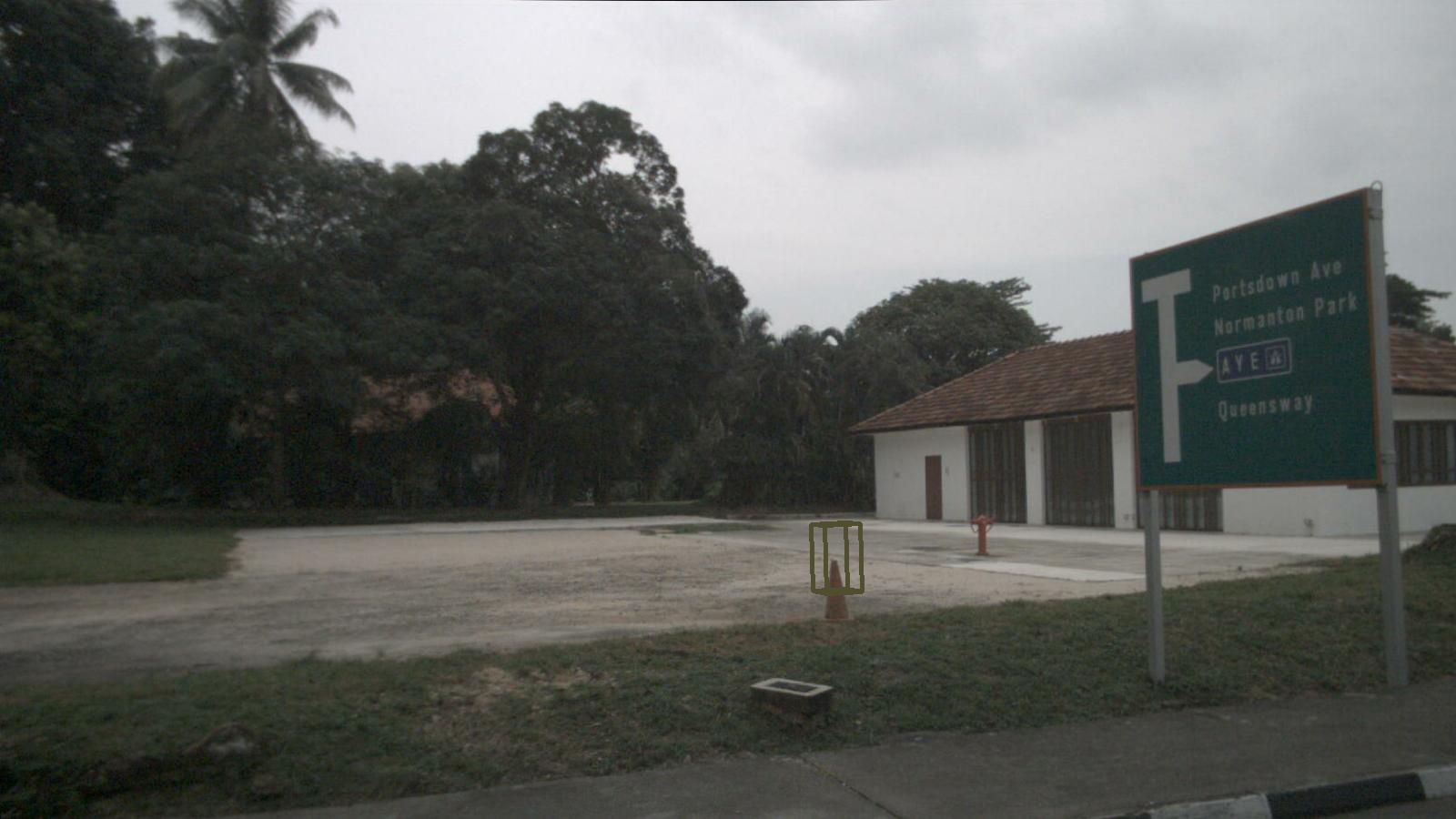}
        \caption{Back Right}
        \label{subfig:cam_back_right_more}
    \end{subfigure}

    \vspace{8pt}
    \begin{subfigure}{0.33\textwidth}
        \centering
        \includegraphics[width=\linewidth]{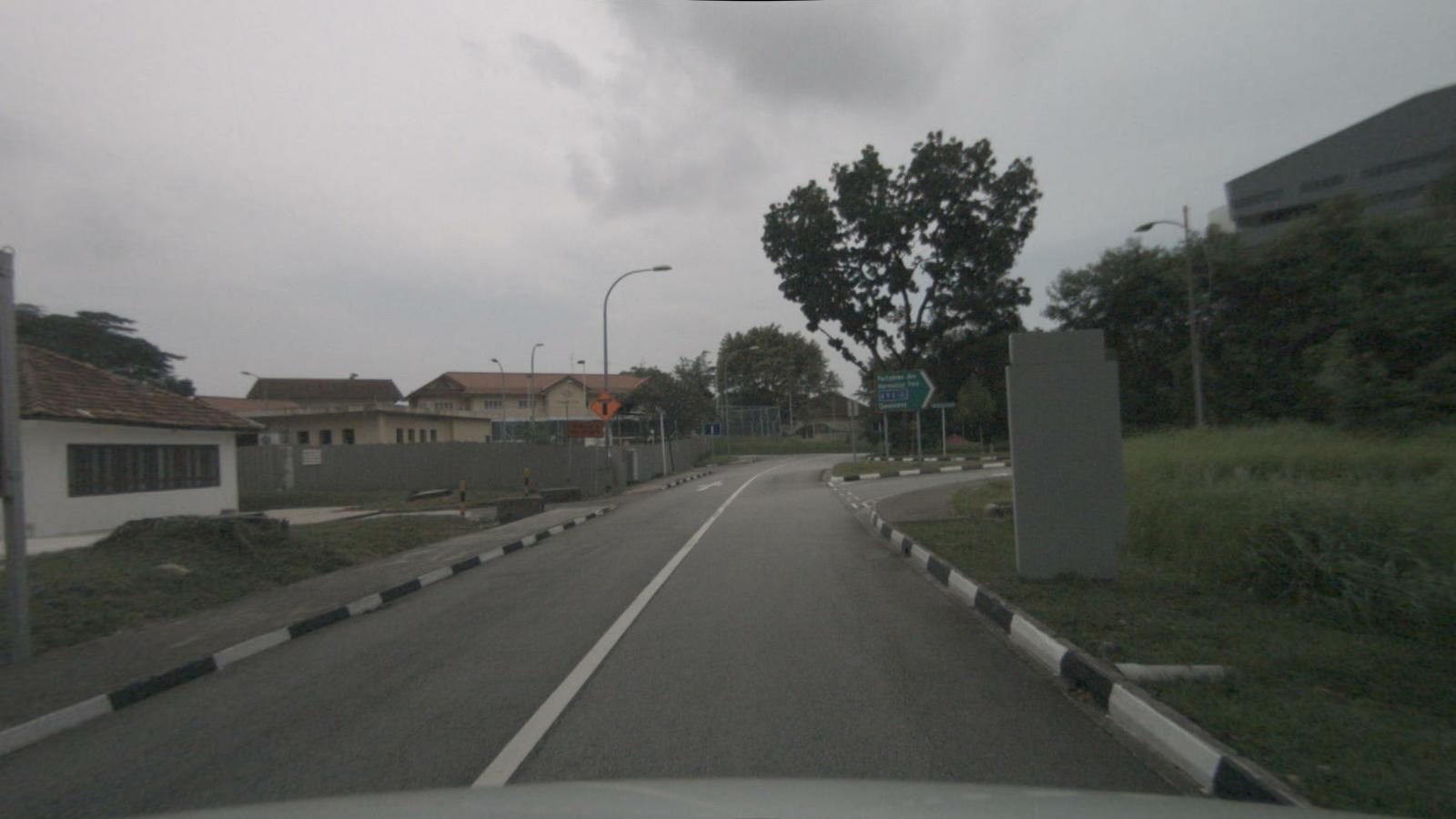}
        \caption{Back}
        \label{subfig:cam_back_more}
    \end{subfigure}
    \hfill
    \begin{subfigure}{0.33\textwidth}
        \centering
        \includegraphics[width=\linewidth]{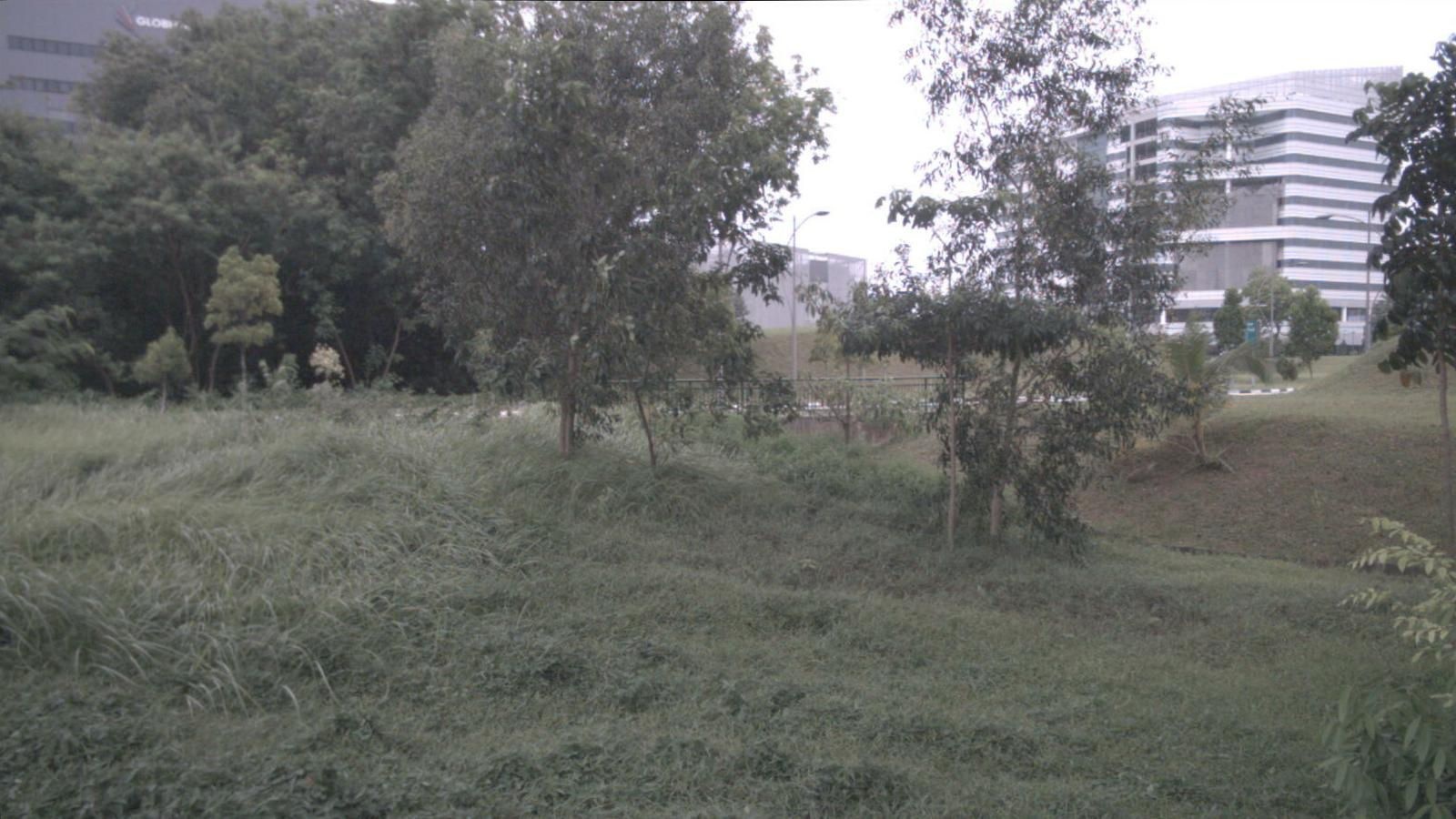}
        \caption{Back Left}
        \label{subfig:cam_back_left_more}
    \end{subfigure}
    \hfill
    \begin{subfigure}{0.33\textwidth}
        \centering
        \includegraphics[width=\linewidth]{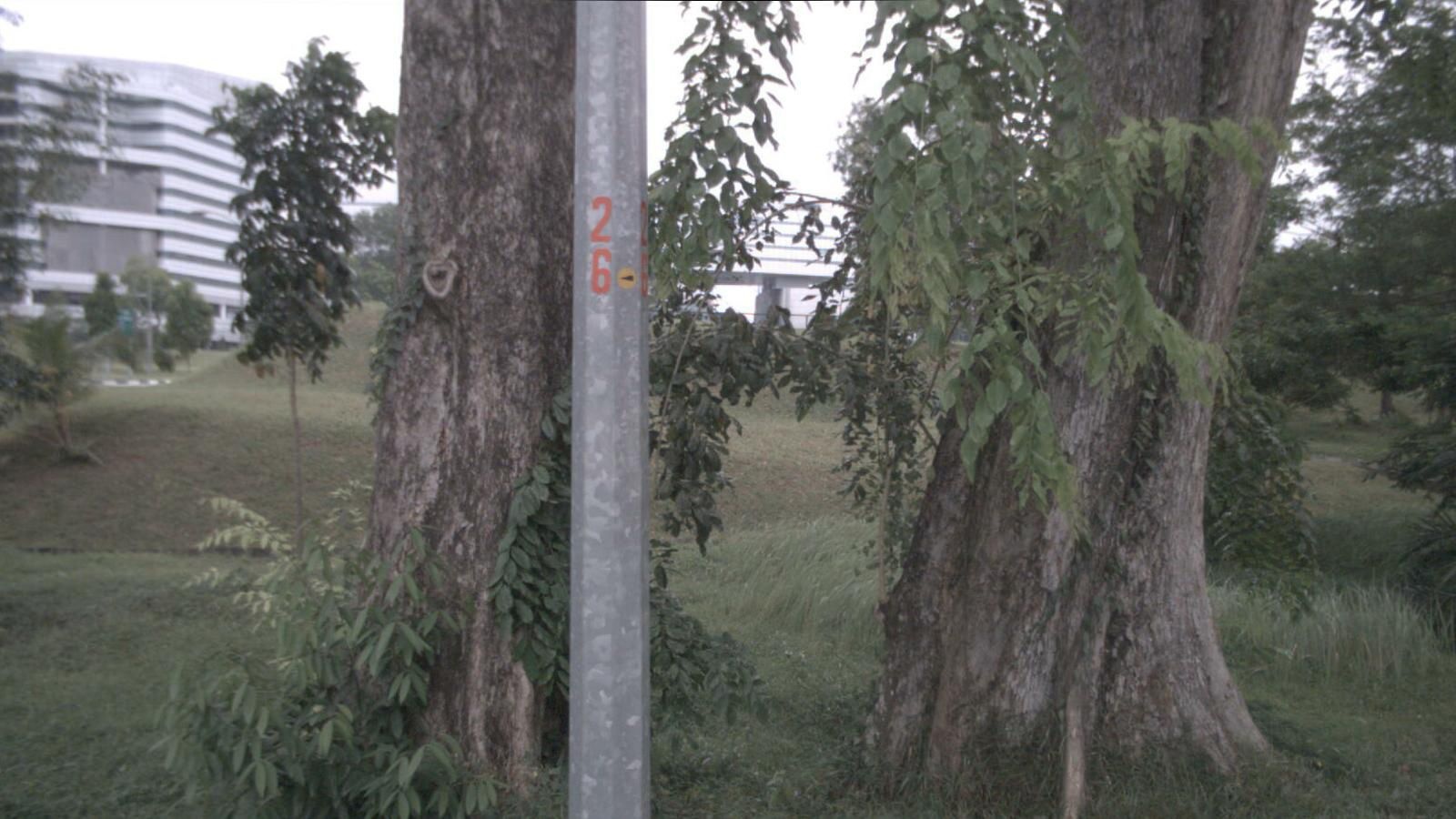}
        \caption{Front Left}
        \label{subfig:cam_front_left_more}
    \end{subfigure}

    \caption{Visualization of multi-view images from the nuScenes dataset, which intuitively illustrates the reliable long-range detection performance of the proposed model on relatively dense traffic targets, scattered across a representative suburban driving scene.}
    \label{fig:nuscenes_cameras_more}
\end{figure*}

\subsection{Ablation Study}
We conduct an ablation study to disentangle the contributions of our key components, focusing on the decomposed operator-native pipeline and enhanced depth-aware feature representation.

\noindent \textbf{Backbone Upscaling.} When scaling the backbone, our Fast-BEV++ (R101) attains an mAP of 0.414, outperforming the Fast-BEV (R101) baseline. This result demonstrates that our performance improvements do not come at the cost of detection accuracy, confirming that our architectural refinements effectively refine feature representation and boost 3D object detection precision. In general, prioritizing the tractability of deployment catalyzes the development of more robust and high-performance perception systems.

\noindent \textbf{Depth Supervision.} We evaluate depth supervision (D) by comparing pure-vision (C) and depth-supervised (C\&D) settings. Incorporating depth supervision consistently improves performance across backbone architectures, yielding notable gains in detection accuracy and perception metrics. This demonstrates depth supervision's effectiveness in enhancing framework robustness.

\subsection{Discussion}
\label{sec:discussion}

Building upon the quantitative experimental results, this section examines the core robustness–efficiency trade-off inherent in our strict one-to-one voxel-pixel mapping, clarifying why this lightweight simplification preserves overall competitive perception performance.

At the core of our design, assigning each voxel to a single camera view eliminates concurrent hardware-level memory write conflicts and enables a standard, conflict-free \textit{Index-Gather-Reshape} pipeline. By leveraging deterministic offline pre-sorting, we achieve a contiguous memory layout that ensures near-deterministic and predictable inference latency. Admittedly, this discards multi-view redundancy in overlapping regions, raising potential concerns regarding unexpected single-camera failures. However, this is mitigated by the inherent structural regularities of multi-camera projective geometry. Because overlapping voxels are resolved deterministically (e.g., via fixed index priority), they are sparsely interleaved across boundaries. A localized camera malfunction yields only isolated blind spots rather than contiguous regions. Consequently, the subsequent BEV encoder effectively interpolates these gaps by aggregating rich local spatial and semantic contexts, preserving the integrity of the BEV feature maps.

Crucially, experiments demonstrate that the combination of learnable depth-aware fusion and the \textit{strict one-to-one mapping} maintains or even improves the detection accuracy. We attribute this favorable performance to the following two factors:
\begin{itemize}
    \item \textbf{Geometric Sparsity:} Overlapping frustums occupy only a marginal fraction of the entire surrounding-view space. For most voxels, feature extraction is inherently single-camera, meaning the overall impact of discarding multi-view aggregation is practically negligible.
    \item \textbf{Semantic-Geometric Continuity:} High-level feature representations exhibit strong spatial coherence and semantic smoothness. This continuity enables robust detection despite localized missing information, as the model excels at contextual aggregation and benefits from the precise geometric guidance provided by additional depth module.
\end{itemize}

In summary, our design avoids the computational overhead of complex multi-view aggregation while delivering highly competitive accuracy, providing a fast and deterministic runtime essential for practical real-time autonomous systems.

\section{Conclusion}
\label{sec:conclusion}

Fast-BEV++ effectively overcomes the trade-off between perception performance and deployment feasibility. Our framework demonstrates that a deployment-aware design philosophy catalyzes \textit{state-of-the-art} performance rather than limiting it. By reframing view transformation as a standard \textit{Index-Gather-Reshape} pipeline, Fast-BEV++ achieves unprecedented deployment efficiency without compromising accuracy. This operator-native approach enables seamless deployment across diverse hardware platforms, making it suitable for production autonomous driving.

The decomposed architecture additionally enables efficient depth-aware fusion, effectively enhancing perception accuracy. Empirical validation on nuScenes establishes a new accuracy benchmark while maintaining real-time performance on production-level hardware. Fast-BEV++ represents a novel approach based on memory-efficient design, demonstrating that high performance and deployment efficiency can seamlessly coexist under the guiding principle of being \textit{Fast by Algorithm, Deployable by Design}.

\vfill\pagebreak

\bibliographystyle{IEEEbib}
\bibliography{strings,refs}

\end{document}